\documentclass[sigconf]{acmart}
\usepackage{hyperref}
\AtBeginDocument{%
  }

\setcopyright{acmlicensed}
\copyrightyear{2025}
\acmYear{2025}
\acmDOI{XXXXXXX.XXXXXXX}
\acmISBN{978-1-4503-XXXX-X/2018/06}

\renewcommand\footnotetextcopyrightpermission[1]{}
\begin{document}

% self-defined package
% \usepackage{multirow}
% \usepackage{arydshln}

% prev title from chenrui
% \title{Fine-grained Question Answering for Practical AI-Generated Video Quality Assessment}
% prev title from tangjing
% \title{FingER: Fine-grained Evaluator with Reasoning for Practical AI-Generated Video Quality Assessment}
% prev title from chenrui
% \title{FingER: Fine-grained Entity-aware Reasoning Model for Practical AI-Generated Video Quality Assessment}
% prev title from sunlei
\title{FingER: Content Aware Fine-grained Evaluation with Reasoning for AI-Generated Videos}
% \title{FingER: Fine-grained Evaluation with Entity-level Reasoning for AI-generated Video Quality Assessment}


\author{Rui Chen}
\email{chenrui.chen@alibaba-inc.com}
\affiliation{%
  \institution{AMAP, Alibaba Group}
  \city{Beijing}
  \country{China}
}

\author{Lei Sun}
\email{ally.sl@alibaba-inc.com}
\affiliation{%
  \institution{AMAP, Alibaba Group}
  \city{Beijing}
  \country{China}
}

\author{Jing Tang}
\email{guangyu.tj@alibaba-inc.com}
\affiliation{%
  \institution{AMAP, Alibaba Group}
  \city{Beijing}
  \country{China}
}

\author{Geng Li}
\email{xiaofeng.lg@alibaba-inc.com}
\affiliation{%
  \institution{AMAP, Alibaba Group}
  \city{Beijing}
  \country{China}
}

\author{Xiangxiang Chu}
\email{chuxiangxiang.cxx@alibaba-inc.com}
\affiliation{%
  \institution{AMAP, Alibaba Group}
  \city{Beijing}
  \country{China}
}

\begin{abstract}
Recent advances in video generation have posed great challenges in the assessment of AI-generated content, particularly with the emergence of increasingly sophisticated models. The various inconsistencies and defects observed in such videos are inherently complex, making overall scoring notoriously difficult. In this paper, we emphasize the critical importance of integrating fine-grained reasoning into video evaluation, and we propose \textbf{F}ing\textbf{ER}, a novel entity-level reasoning evaluation framework that first automatically generates \textbf{F}ine-grained \textbf{E}ntity-level questions, and then answers those questions by a \textbf{R}easoning model with scores, which can be subsequently weighted summed to an overall score for different applications. Specifically, we leverage LLMs to derive entity-level questions across five distinct perspectives, which (i) often focus on some specific entities of the content, thereby making answering or scoring much easier by MLLMs, and (ii) are more interpretable. Then we construct a FingER dataset, consisting of approximately 3.3k videos and corresponding 60k fine-grained QA annotations, each with detailed reasons. Based on that, we further investigate various training protocols to best incentivize the reasoning capability of MLLMs for correct answer prediction. Extensive experiments demonstrate that a reasoning model trained using Group Relative Policy Optimization (GRPO) with a cold-start strategy achieves the best performance. Notably, our model surpasses existing methods by a relative margin of 11.8\% on GenAI-Bench and 5.5\% on MonetBench with only 3.3k training videos, which is at most one-tenth of the training samples utilized by other methods. Our code and dataset will be released soon.

\end{abstract}

\maketitle

\begin{figure}[h]
  \centering
  \includegraphics[width=\linewidth]{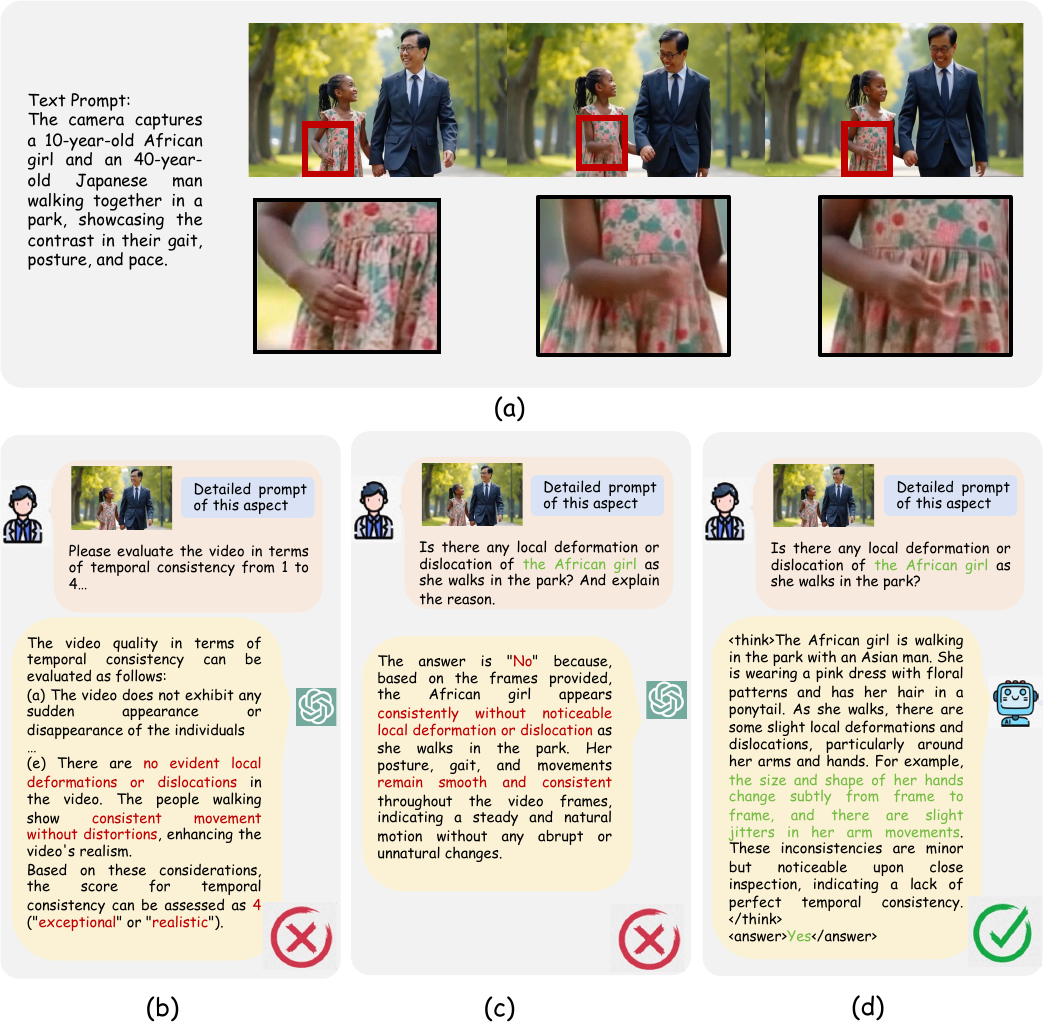}
  \caption{Advanced generation models often exhibit localized defects while maintaining overall visually appealing, as illustrated in (a), which requires fine-grained in-depth understanding. (b) and (c) show that even with detailed instructional prompts and entity-level questions, GPT-4o still fails to identify this hand deformation. (d) shows the effectiveness of our work by integrating reasoning model with fine-grained entity-level questions.}
    \label{fig:example}
  \Description{}
\end{figure}

\section{Introduction}

Recent advancements in Text-to-Video (T2V) generative models \cite{brooks2024video,zheng2024open,bao2024vidu} have demonstrated significant progress in producing visually appealing and content-rich videos. For instance, post-Sora models such as Kling have shown the ability to generate high-resolution videos that closely adhere to textual prompts. However, these models often produce localized artifacts, inconsistencies, and violations of physical laws. These issues highlight the necessity for the development of robust and reliable quality assessment methods for AI-generated video content.

% 统一一下称谓 video quality assessment
%The recent literature in quality assessment of the video generation models has drawn much attention from the academia.
%Previous quality assessment methods can broadly be categorized into three types:

%The recent literature in quality assessment of the video generation models has drawn much attention from the academia. 

Early research on evaluating AI-generated videos has primarily relied on feature-based metrics, such as the Frechet Video Distance (FVD) \cite{unterthiner2019fvd} and optical flow-based methods like RAFT \cite{teed2020raft}. While these methods effectively assess overall visual quality and dynamic characteristics, they fall short in capturing nuanced aspects that require deeper semantic understanding and fine-grained reasoning.
To address these limitations, recent studies have introduced MLLMs for more comprehensive evaluations. For example, VideoScore \cite{he2024videoscore} proposes a framework that evaluates five distinct aspects of video quality using an MLLM to assign scores ranging from 1 to 4. VisionReward \cite{xu2024visionreward} aligns video generation with human perception by formulating predefined judgment questions and fine-tuning a video-based MLLM to compute weighted scores. Similarly, LiFT \cite{wang2024lift} learns a reward model that provides reasons and scores across multiple aspects to align the generation model with human preferences. Despite these advancements, two key challenges persist:

(i) \textbf{Inadequacy of Fine-grained Video Reasoning}: Although advanced generative models have significantly improved global visual quality by reducing issues such as blurriness and flickering, they still exhibit localized spatiotemporal inconsistencies, distortions, unnatural artifacts, and violations of physical laws, especially in scenarios involving complex motion or multiple entities. For instance, Fig~\ref{fig:example}(a) shows a video generated by Pixverse that, despite its high overall visual appeal, contains a noticeably deformed hand in a localized area. This example underscores the need for more fine-grained and context-aware reasoning capabilities in video understanding, moving beyond superficial visual pattern recognition to incorporate temporally grounded and semantically rich analysis.
(ii) \textbf{Domain Gap in AI-Generated Videos}: Current state-of-the-art MLLMs struggle to capture the intrinsic characteristics of AI-generated videos, even with well-defined prompts. As illustrated in Fig~\ref{fig:example}(b) and (c), GPT-4o misidentifies the deformed hand in a video and assigns a high score based on misleading explanations. This issue is primarily attributed to a domain gap between the training data used by MLLMs and the unique features of AI-generated videos. In essence, AI-generated videos can deceive MLLMs in certain latent feature spaces. Bridging this gap requires a high-quality dataset of AI-generated videos. Moreover, developing strategies to enhance the generalization of MLLMs to AI-generated videos remains an open challenge.

Inspired by the Question Generation and Answering (QG/A) framework \cite{cho2024davidsonian} and recent reasoning works \cite{zhou2025r1,liu2025visual,chu2025gpg,liu2025understanding} that demonstrate a significant self-emergence of complex cognitive reasoning abilities induced by Deepseek R1 \cite{guo2025deepseek}, we argue that incorporating fine-grained reasoning abilities would significantly enhance the video quality assessment. In this paper, we propose \textbf{F}ing\textbf{ER}, a novel framework that first decomposes the overall evaluation into fine-grained entity-level questions and then answers these questions with corresponding scores by a reasoning model, which is fine-tuned on our high-quality dataset using GRPO with a cold-start initialization. 
Specifically, we employ five distinct aspects as defined in VideoScore \cite{he2024videoscore}, including text-to-video alignment, temporal consistency, factual consistency, dynamic degree, and visual quality. 
By deriving such fine-grained entity-level questions, our framework not only enables the model to explicitly focus on specific characteristics of certain entities, thereby facilitating a more fine-grained understanding, but also enhances interpretability through these structured QA pairs. 

Based on these questions, we prompted several strong MLLMs \cite{hurst2024gpt,team2024gemini} to provide answers. However, we observed that these models struggle to provide correct answers, particularly in aspects like factual consistency. As stated before, we attribute this to the lack of high-quality AI-generated video datasets and the inadequate reasoning capabilities of current models. Therefore, we curated a fine-grained AI-generated video reasoning dataset, \textbf{F}ing\textbf{ER}-\textbf{Instruct}-\textbf{60k}, which consists of 3.3k AI-generated videos sourced from advanced generation models like Kling, Luma, Vidu, PixVerse, CogVideoX \cite{yang2024cogvideox}, etc. For each video, we generate fine-grained questions and annotate them with 'Yes/No'. To ease human labor and also reduce potential errors, we leverage MLLMs to generate detailed reasoning explanations given each question and its answer. (Note that,  while MLLMs often struggle to answer these questions correctly, they demonstrate higher possibilities of producing coherent reasoning when the answer is explicitly provided, suggesting the presence of underlying reasoning capabilities.) These generated reasons were subsequently re-checked and refined by human annotators to ensure accuracy and quality. At last, we collect 60k fine-grained QA annotations with high-quality detailed reasons. 

To enhance the video reasoning capabilities, we choose Qwen2.5-VL \cite{bai2025qwen2}, and explore multiple training protocols on our dataset, including directly training with answers, training with reasons, zero GRPO training and GRPO training with a cold-start initialization. Our experiments reveal that integrating high-quality reasons can largely increase the performance along with the interpretability, and GRPO with cold-start can further enhance its performance, especially in dimensions that require in-depth understanding. We also test our reasoning model in a zero-shot manner on public benchmarks, and still consistently achieve state-of-the-art performance.

In summary, we propose an entity-level quality assessment framework with strong reasoning and generalization capabilities. To the best of our knowledge, our work is the first to introduce entity-level reasoning into the quality assessment of AI-generated videos.

Our contributions can be summarized as follows:
\begin{itemize}
    \item \textbf{Novel Evaluation Approach.} We propose a novel evaluation approach \textbf{F}ing\textbf{ER}, designed for practical AI-generated video quality assessment. It comprises an entity-level question generation module and a video reasoning model that provides corresponding scores. By emphasizing fine-grained reasoning, our approach effectively addresses localized defects in AI-generated videos that require in-depth understanding and significantly enhances interpretability.
    \item \textbf{Fine-grained Reasoning Dataset}. We present a new dataset for AI-generated video reasoning, containing 3.3k videos and 60k entity-level QA annotations sourced from advanced generation models. Each QA pair is annotated with detailed reasons. This dataset aims to further advance research in this field.
    \item \textbf{Enhanced Training Protocols.} 
    We explore multiple training protocols to enhance the fine-grained video reasoning capabilities of MLLMs. Notably, we are the first to introduce GRPO training into AI-generated video quality assessment, which proves to be highly effective in improving both reasoning and generalization abilities
    \item \textbf{Strong Performance}. Extensive experiments demonstrate the effectiveness of our approach. We achieve state-of-the-art performance on public benchmarks using only one-tenth of the training videos, thereby highlighting the superior generalization capability of our model.
\end{itemize}
% 
% ACM's consolidated article template, introduced in 2017, provides a
% consistent \LaTeX\ style for use across ACM publications, and
% incorporates accessibility and metadata-extraction functionality
% necessary for future Digital Library endeavors. Numerous ACM and
% SIG-specific \LaTeX\ templates have been examined, and their unique
% features incorporated into this single new template.
% 
% If you are new to publishing with ACM, this document is a valuable
% guide to the process of preparing your work for publication. If you
% have published with ACM before, this document provides insight and
% instruction into more recent changes to the article template.
% 
% The ``\verb|acmart|'' document class can be used to prepare articles
% for any ACM publication --- conference or journal, and for any stage
% of publication, from review to final ``camera-ready'' copy, to the
% author's own version, with {\itshape very} few changes to the source.

\section{Related Work}
% As noted in the introduction, the ``\verb|acmart|'' document class can
% be used to prepare many different kinds of documentation --- a
% double-anonymous initial submission of a full-length technical paper, a
% two-page SIGGRAPH Emerging Technologies abstract, a ``camera-ready''
% journal article, a SIGCHI Extended Abstract, and more --- all by
% selecting the appropriate {\itshape template style} and {\itshape
%   template parameters}.

% This document will explain the major features of the document
% class. For further information, the {\itshape \LaTeX\ User's Guide} is
% available from
% \url{https://www.acm.org/publications/proceedings-template}.

%\subsection{Text-to-Video Generative Models}

% The primary parameter given to the ``\verb|acmart|'' document class is
% the {\itshape template style} which corresponds to the kind of publication
% or SIG publishing the work. This parameter is enclosed in square
% brackets and is a part of the {\verb|documentclass|} command:
% \begin{verbatim}
%   \documentclass[STYLE]{acmart}
% \end{verbatim}

% Journals use one of three template styles. All but three ACM journals
% use the {\verb|acmsmall|} template style:
% \begin{itemize}
% \item {\texttt{acmsmall}}: The default journal template style.
% \item {\texttt{acmlarge}}: Used by JOCCH and TAP.
% \item {\texttt{acmtog}}: Used by TOG.
% \end{itemize}

% The majority of conference proceedings documentation will use the {\verb|acmconf|} template style.
% \begin{itemize}
% \item {\texttt{sigconf}}: The default proceedings template style.
% \item{\texttt{sigchi}}: Used for SIGCHI conference articles.
% \item{\texttt{sigplan}}: Used for SIGPLAN conference articles.
% \end{itemize}

\begin{figure*}[t]
  \includegraphics[width=\linewidth]{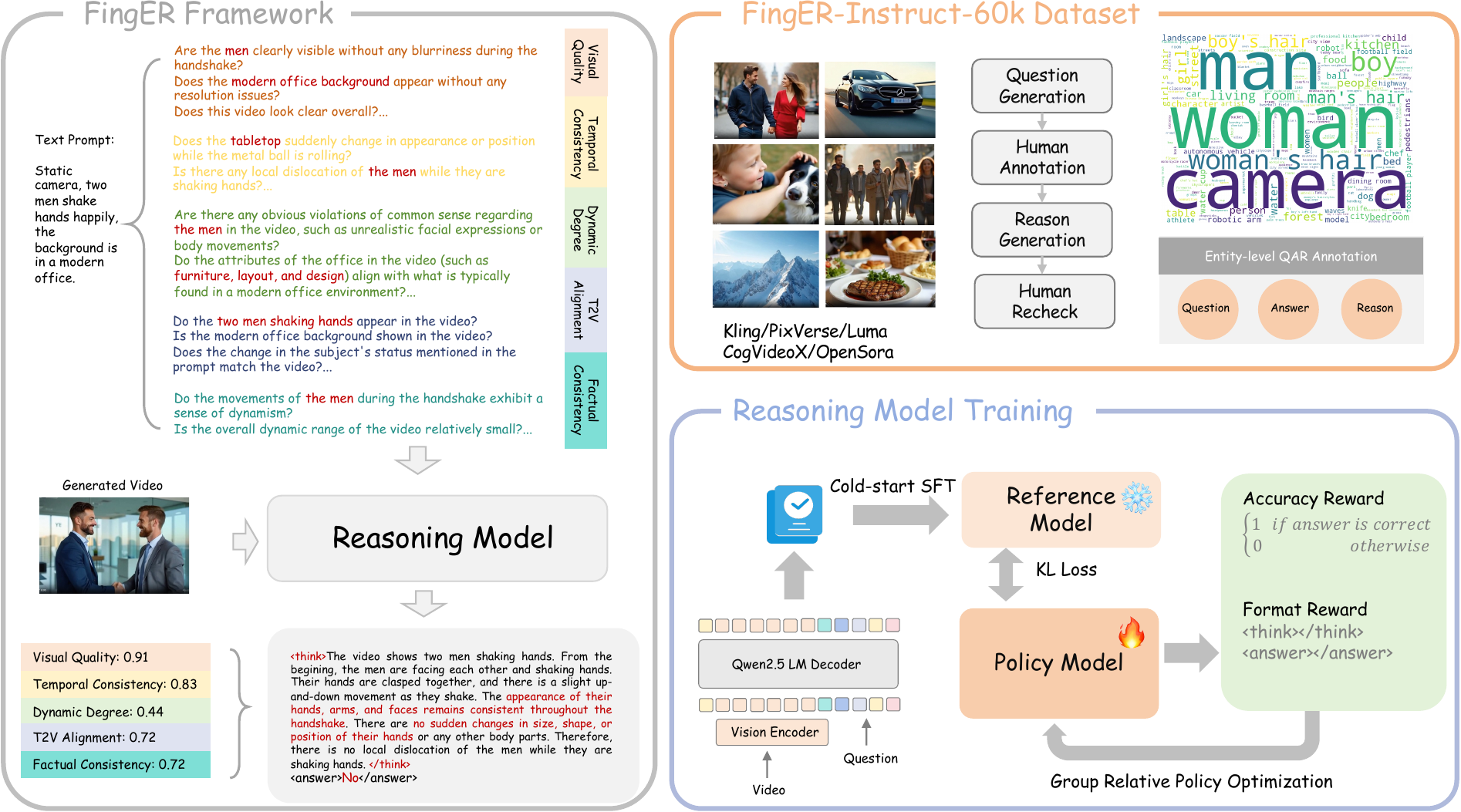}
  \caption{The overview of our proposed FingER framework, including (a) the evaluation pipeline, (b) FingER-Instruct-60k dataset curation, and (c) GRPO training of our reasoning model.}
  \Description{main figure}
  \label{fig:pipeline}
\end{figure*}

% \subsection{Text-to-Video Evaluation / Video Quality Assessment}
\subsection{Video Quality Assessment}

Early approaches relied on feature-based metrics, such as Fréchet Video Distance (FVD) \cite{unterthiner2019fvd}, Inception Score (IS) \cite{is}, and CLIPSim \cite{clipsim}. And benchmark works like EvalCrafter \cite{evalcrafter2024} and VBench \cite{vbench2023} introduced comprehensive evaluation frameworks with 18 and 16 metrics, respectively. However, these methods fall short in assessing deep semantic understanding or aligning with human perception.

With the rapid advancement of MLLMs \cite{bai2025qwen2,team2024gemini,chen2024internvl,chu2024mobilevlm}, increasing studies have explored to leverage their capabilities to facilitate image/video quality evaluation \cite{wu2024q,cho2024davidsonian,li2025next,lin2024evaluating,huang2024mmgenbench}. Inspired by DSG \cite{cho2024davidsonian}, which uses question generation/answering (QG/A) for interpretable assessment, T2VScore \cite{t2vscore} adopted a QA framework for T2V alignment. T2VQA \cite{t2vqa} introduced the T2VQA-DB dataset, comprising 10k videos annotated with Mean Opinion Scores (MOS), and trained a transformer-based model to predict these scores. Similarly, VideoScore \cite{he2024videoscore} proposed a larger dataset across five dimensions and employed a MLLM for scoring.
VMBench \cite{ling2025vmbench} introduced perception-aligned motion metrics to evaluate motion quality.
While these methods predict scores or labels, they often overlook the reasoning behind assessments, limiting their effectiveness.
Our work distinguishes itself by incorporating entity-level reasoning for evaluating advanced generation models reliably. 

Another line of research focuses on reward models for improving generative models via Reinforcement Learning from Human Feedback (RLHF), such as Diffusion-DPO \cite{wallace2024diffusion}, VisionReward \cite{xu2024visionreward} and UnifiedReward \cite{wang2025unified}. While these efforts target generative model optimization, our work emphasizes practical video quality evaluation, we expect it is able to further benefit the generation models using RLHF in future work.

% In addition to specifying the {\itshape template style} to be used in
% formatting your work, there are a number of {\itshape template parameters}
% which modify some part of the applied template style. A complete list
% of these parameters can be found in the {\itshape \LaTeX\ User's Guide.}

% Frequently-used parameters, or combinations of parameters, include:
% \begin{itemize}
% \item {\texttt{anonymous,review}}: Suitable for a ``double-anonymous''
%   conference submission. Anonymizes the work and includes line
%   numbers. Use with the \texttt{\string\acmSubmissionID} command to print the
%   submission's unique ID on each page of the work.
% \item{\texttt{authorversion}}: Produces a version of the work suitable
%   for posting by the author.
% \item{\texttt{screen}}: Produces colored hyperlinks.
% \end{itemize}

% This document uses the following string as the first command in the
% source file:
% \begin{verbatim}
% \documentclass[sigconf,authordraft]{acmart}
% \end{verbatim}

\subsection{Reasoning Inference in Large Models}
Reasoning inference aims to emulate human-like thinking processes by forming the final answer through a Large Language Model (LLM). Specifically, to answer a given question, an LLM is required to think divergently and record the thinking processes, which are subsequently referenced when formulating the final answer. This approach has inspired a variety of research, including prompting-based Chain-of-Thought (CoT) \cite{wei2022chain}, planning-based Graph-of-Thought \cite{besta2024graph} and Tree-of-Thought \cite{yao2023tree} processing, reward methods \cite{lai2024step}, and supervised fine-tuning (SFT) datasets with sufficient context \cite{ye2025limo}. Notably, DeepSeek-R1 \cite{guo2025deepseek} integrates specific prompts with reinforcement learning (RL), enabling the model to first generate the thinking process before producing the final answer. This method allows for supervised fine-tuning with a small amount of annotated data containing thinking processes, followed by reinforcement learning fine-tuning on more data without thinking processes. A very recent approach \cite{chu2025gpg} proposes a highly simplified reinforcement learning framework and demonstrates its validity across several benchmarks.
% However,  reasoning  Multi-modal Large-Language-Models (MLLMs) needs more explorations.
% GRPO

% \section{Method / FingER}
\section{Method}
% 
% \subsection{Overview}
% 
In this section, we first introduce our entity-level video quality assessment framework --- \textbf{F}ing\textbf{ER} in Sec.~\ref{subsec:method_framework}. 
% which consists of three parts: (i) multi-dimensional entity-level question generation, (ii) reasoning model for question answering with detailed thinking process, and (iii) scoring function that converts natural language responses to quantative scores, as illustrated in Fig.~\ref{fig:pipeline}.
% --- --- --- for short --- --- ---
% [TODO] to demonstrate the entire quality assessment pipeline. 
Then, we detail the data curation pipeline of our proposed dataset, namely \textbf{F}ing\textbf{ER}-\textbf{Instruct}-\textbf{60k} in Sec.~\ref{subsec:method_dataset}. 
% instruction tuning 
% We detail the dataset construction process with (i) prompt and Text-to-Video (T2V) model selection, (ii) entity-aware question generation and annotation, and (iii) reasoning generation and verification.
% --- --- --- for short --- --- ---
% answering (QG/A) to reasoning process generation. 
% In Sec.~\ref{subsec:method_training}, 
In the end, we combine multiple training methods with our proposed instruction tuning dataset, from the basic supervised fine-tuning (SFT), to reasoning training with reinforcement learning (RL), as detailed in Sec.~\ref{subsec:method_training}. 
% 
% from the basic supervised fine-tuning (SFT) paradigm, to SFT combined with reasoning process, and utilize reinforcement learning (RL) with group relative policy optimization  to further improve model's logical thinking capabilities.
% --- --- --- for short --- --- ---
% 
% \begin{figure*}
%   \includegraphics[width=\linewidth]{samples/figures/test_new_sunlei-crop_main.pdf}
%   \caption{Demonstration of our fine-grained evaluation results}
%   \Description{}
% \end{figure*}
% 
% overall introduction
% \subsection{FingER: Entity-aware VQA Framework}

\subsection{Entity-level VQA Framework}
\label{subsec:method_framework}

% entity-extraction
For Text-to-Video (T2V) generation task, user input prompt is the only key instruction for generative models to understand and generate content that well-aligned with user's intent. 
To perform entity-level quality assessment of AI-generated videos, we start from understanding the user's prompt through extracting entities, attributes, and actions within itself. 
Inspired by DSG \cite{cho2024davidsonian} in Text-to-Image (T2I) evaluation, we also utilize closed-source Large-Language-Model (LLM) to perform textual understanding and the following entity extraction.
% to extract entities from text prompt.
As shown in Fig.~\ref{fig:pipeline}, we provide abundant in-context learning (ICL) \cite{wei2022chain} examples from different video generation scenarios and formulate the final input for GPT-4o \cite{hurst2024gpt}, in which way we can harvest more steady entity extraction results.
% formulate the input prompt for GPT-4o\cite{}.
% The intuition behind entity-aware question generation is that we hope to guide the MLLM to focus on understanding the correlation between entity-level textual description and the corresponding visual appearance based on the video content
% 
% \textbf{[TODO]} explain why entity-level, our intuition behind that

With entities extracted from the user's prompt, we generate entity-level questions from \textbf{five} distinct video quality assessment dimensions,
% for the AI-generated videos, 
including \textit{visual quality}, \textit{text-to-video alignment}, \textit{temporal consistency}, \textit{factual consistency}, and \textit{dynamic degree}. 
For each dimension, we provide a detailed explanation followed by several key points, formulating the context information when prompting the LLM. 
% In order to help the LLM better understand which question should be asked when coping with a specific entity, 
% Similarly, 
We also prepare adequate \textbf{entity-level} in-context learning examples, which are summarized from videos with and without obvious artifacts or hallucinations.
In this way, we can help the LLMs to better understand which question should be asked when coping with a specific entity along with the given assessment dimension. 
In short, we break down the granularity of fine-grained video quality assessment from multi-dimensional level to entity-level.
And the intuition behind entity-level question generation is that we hope fine-grained question/answering can guide the MLLM to focus on understanding the correlation between entity-level textual description and its corresponding visual appearance based on the video content.

After the entity-level question generation procedure, our fine-tuned MLLM answers the above questions with a simple \textit{"Yes"} or \textit{"No"}, along with a detailed reasoning process explaining why the answer is that. 
Learning the logical reasoning process is critical for model performance improvements, as detailed in the experiment Sec.~\ref{exp:sft_rl}. The outputted reason can also be useful when conducting practical video quality assessment, which is more interpretable and user-friendly.
% 
% 解释由 entity 汇总到 dimension，由 dimension 汇总到整个视频的计算得分的过程
% \textbf{[TODO]} why reasoning?, explain why we use probability instead of simple Yes / No
% 
To formulate a final score representing the overall quality of AI-generated videos, we start by calculating the probability of the answer token ("Yes" or "No") for each entity-level question to represent the \textbf{entity-level score}.
% in the first place. 
% Following the next token prediction paradigm, 
% We first gather the token set for "Yes" and "No", 
Since there are multiple "Yes" and "No" with different formats but similar meanings in the vocabulary of our MLLM, we first gather the token set for "Yes" and "No". 
In this paper, we take $\left[ "Yes", "yes", "YES", ""Yes", "\, Yes" \right]$ as the token set for answer "Yes", 
and $\left[ "No", "no", "NO", ""No", "\, No" \right]$ for answer "No", denoted by $T_{Y}$ and $T_{N}$, respectively.
With logits from the answer token, we extract all the logit whose token id is within the token set, and apply softmax over $T_{Y} \cup T_{N}$, 
as illustrated in Eq.~\ref{eq:entity_prob}. Then, given the entity-level question $q$, we can get the answer's probability $P(No \, | \, q)$ and $P(Yes \, | \, q)$ with a simple sum up.
% 
% \textbf{[TODO]} softmax problem.
% 
\begin{equation}
\label{eq:entity_prob}
    \begin{aligned}
        & P(No \; | \; q) = \sum_{i = 1}^{n}Softmax(x_{i}) \, , \; x_{i} \in T_{N} \, ;\\
        & P(Yes \; | \; q) = \sum_{j = 1}^{m}Softmax(y_{j}) \, , \; y_{j} \in T_{Y} \, .
    \end{aligned}
\end{equation}

Instead of directly using the derived probability as the entity-level score, we still need the judgment on whether the question is positive or negative. 
% with GPT-4o\cite{hurst2024gpt}. 
% \textbf{[TODO] use factual consistency question as example}
% 
% For example, given the question \textit{"Is the metal ball clearly visible throughout its movement on the tabletop?"} 
For example, given the question \textit{"Do the attributes of the table in the video (such as size, shape, and material) align with real-world characteristics?"} from the factual consistency dimension, it is apparent that the factual consistency of the assessed video goes up with a positive \textit{"Yes"} answer.
% \textit{"Yes"}. 
We define this type of question as a positive one, and vice versa. 
We denote the status of an entity-level question with $q_{stat}$, if $q_{stat}$ equals 1, it means that the question is positive; otherwise, the question is negative. 
\begin{equation}
\label{eq:entity_score}
    S_{entity} = \left\{
    \begin{aligned}
        & P \, (No \; | \; q), \; & if \; q_{stat} \, = \, 0 \, ; \\
        & P \, (Yes \; | \; q), \; & if \; q_{stat} \, = \, 1 \, .
    \end{aligned}
    \right. 
\end{equation}
With the aforementioned preparations setup, we propose our entity-level score $S_{entity}$, which correlates positively with the quality of the assessed video. 
When the entity-level question is positive, we use the probability of the "Yes" answer $P(Yes \, | \, q)$ to represent the score it can gain. 
And we utilize the probability of the "No" answer $P(No \, | \, q)$ if the question is negative, as illustrated in Eq.~\ref{eq:entity_score}. 
In short, our intuition behind this design is that as long as the video quality goes up with which answer, we calculate our entity-level score based on that answer's probability.
% if the answer is \textit{"Yes"}, then the entity-level score $S_{entity}$ should reflect the gains with it, 
% denoted by $q_{stat}$ \textbf{[TODO]} explain variable.
% \textbf{[TODO]} explain variable.
% 
% \begin{equation}
% \label{eq:entity_score}
%     S_{entity} = \left\{
%     \begin{aligned}
%         & P \, (No \; | \; q), \; & if \; q_{stat} \, = \, 0 \, ; \\
%         & P \, (Yes \; | \; q), \; & if \; q_{stat} \, = \, 1 \, .
%     \end{aligned}
%     \right. 
% \end{equation}
% 
% \begin{equation}
%     P(Yes \; | \; q) = \sum_{i = 1}^{n}\frac{e^{x_{i}}}{\sum_{i = 1}^{n}e^{x_{i}} + \sum_{j = 1}^{m}e^{y_{j}}}
% \end{equation}
% 
% 
% \textbf{[TODO]} finish $S_{dim}$ and $S_{overall}$
% 
Then, we utilize entity-level question/answering pairs that are under the same quality assessment dimension to formulate our \textbf{dimension-level score} $S_{dim}$. 
To be specific, we simply calculate the linear summation of multiple answers' probability $S_{entity}$, as illustrated in Eq.~\ref{eq:score_dim}.
% which are from the corresponding entity-aware questions that are under the same quality assessment dimension.
\begin{equation}
\label{eq:score_dim}
    S_{dim} = \sum_{i = 1}^{N}S_{entity}i
\end{equation}
In the end, we derive the \textbf{overall-level score} $S_{overall}$ with the weighted average of five distinct dimension scores $S_{dim}$ in Eq.~\ref{eq:score_overall}.
\begin{equation}
\label{eq:score_overall}
    S_{overall} = \sum_{i = 1}^{5}w_{i} \, . \, S_{dim}i
\end{equation}
% 
% to represent the overall score
% judgement questions
% For each dimension, for example, 
% 
In short, we propose the entity-level VQA framework \textbf{F}ing\textbf{ER}, which consists of three parts: (i) entity-level question generation, (ii) the fine-tuned MLLM with reasoning output, and (iii) the hierarchical scoring function that converts token probability to multi-level scores. 

% To the best of our knowledge, we are the first to propose the entity-aware video quality assessment framework, named \textbf{F}ing\textbf{ER}, 
% which consists of three parts: (i) entity extraction and entity-aware question generation, (ii) fine-tuned MLLM for question answering with detailed reasoning process, and (iii) the hierarchical scoring function that converts token probability to entity-level score, and formulates dimension-level and overall-level score with 
% simply linear summation and weighted average. 
% --- --- --- for short --- --- ---
% \textbf{[TODO]} conclusion of 1st part.
% we start the question/answering generation process.
% question generation
% reasoning model
% Modifying the template --- including but not limited to: adjusting
% margins, typeface sizes, line spacing, paragraph and list definitions,
% and the use of the \verb|\vspace| command to manually adjust the
% vertical spacing between elements of your work --- is not allowed.

% {\bfseries Your document will be returned to you for revision if
%   modifications are discovered.}

% \subsection{FingER-Instruct: Entity-level QG/A Dataset}
% \subsection{FingER-Instruct: Entity-level QG/A Dataset with Reasoning}
\subsection{Entity-level Dataset with Reasoning}
\label{subsec:method_dataset}
In this section, we introduce the construction pipeline of our entity-level instruction tuning dataset, named \textbf{F}ing\textbf{ER}-\textbf{Instruct}-\textbf{60k}.
% 
% Our data construction pipeline mainly consists of three parts: 
% in the first part, we begin with the prompt and Text-to-Video (T2V) model selection, in which we select around 3.3k videos generated by 8 modern T2V models with 420 diverse text prompts, as detailed in Sec.~\ref{dataset:selection}; 
% the second part mainly focuses on how we generate the entity-level question in detail, along with the annotation process of its corresponding answer, referred in Sec.~\ref{dataset:qa}; 
% and the third part (in Sec.~\ref{dataset:reasoning}) is about how we properly prompt the closed-source MLLM \cite{team2024gemini} to generate reasonable explanation for each entity-level question, accompanied by the human-in-the-loop process that fixes the inappropriate reasoning process.
% --- --- --- for short --- --- ---
% 
\subsubsection{\textbf{Prompt and T2V Model Selection}}
\label{dataset:selection}
Based on VideoGen-Eval \cite{zeng2024dawn} dataset, our instruction tuning dataset is composed of 420 diverse text prompts and 3.3k AI-generated videos produced by 8 modern T2V models, including closed-source models: Kling, Luma, PixVerse, Vidu, Qingying, and open-sourced models: Mochi-1 \cite{genmo2024mochi}, CogVideoX \cite{yang2024cogvideox}, Open-Sora \cite{zheng2024open}. 
% 
% All the generated videos are from VideoGen-Eval\cite{zeng2024dawn}.
% 
% 
We utilize all 420 text prompts from the T2V session \cite{zheng2024open}, which cover a diverse range of complex scenarios, including human-centric activities, material and spatial relationships, as well as animal and text generations. These prompts are derived from real-life user inputs.
% As for the T2V models, 
% we select the generative models uniformly based on the quality of their generated videos.
% quality
% assessment perspectives, from generating excellent quality videos to average quality videos.
As for the T2V model selection, we denote models that understand and obey most of the common sense and physical laws, and generate time-consistent videos without obvious temporal distortions as the high-quality model. 
We select the generative models uniformly based solely on the quality of their generated videos, spanning from high-quality models to average-quality models, for a more diverse training data distribution. 
% 
% We manually categorize the video generation models into three levels from the video quality assessment perspectives: (i) high-level models, which understand and obey most of the common sense and physical laws, generate more time-consistent videos without obvious temporal distortion; (ii) medium-level videos, 

\subsubsection{\textbf{Entity-level Question Generation and Annotation}}
\label{dataset:qa}
Our multi-dimensional entity-level question generation starts with understanding users' input prompts and extracting the entities within.
We use GPT-4o \cite{hurst2024gpt} for prompt understanding and entity extraction, with abundant in-context learning examples provided.
Then, we perform the entity-level question generation for our five distinct assessment dimensions.
For each entity, we prompt the LLM with task introduction, assessment dimension explanation with several key points to focus on, user's input prompt, the extracted entity, and the most important in-context learning examples. And we extract the generated questions with regular expression matching.

For data annotation, we engaged 10 professional annotators to complete the task of annotating 60k question/answer pairs. Inter-annotator agreement was ensured through multiple rounds of small-scale pilot annotations, and the entire process took approximately one month to complete.
% 
% T2V models and QA annotation

\subsubsection{\textbf{Reasoning Generation and Verification}}
\label{dataset:reasoning}
We employ the powerful closed-source MLLM \cite{team2024gemini} to generate the initial version of the reasoning process. Specifically, we prompt the MLLM with the assessment dimension explanation, user prompt, in-context learning examples, and the entity-level question along with its human-annotated result. An interesting finding is that when the MLLM is provided with the correct answer to the entity-level question, the generated reasoning process for explaining the answer is more reasonable than when directly generating the answer and its reason. Rather than using the MLLM-generated reasoning process directly, we conduct thorough human verification to ensure the quality of our reasoning training data.
% asking for the reason of the solely question.

% \textbf{[TODO]} train / test split
With aforementioned entity-level questions, human-annoted answers and detailed reasoning process, we formulate our instruction tuning dataset \textbf{F}ing\textbf{ER}-\textbf{Instruct}-\textbf{60k}, which serves as our basis for the model training in the next section.

% \subsection{Instruction Tuning with Proposed Dataset}
\subsection{Instruction Tuning and GRPO Training}
\label{subsec:method_training}
We use Qwen2.5-VL-7B-Instruct \cite{bai2025qwen2} as our base model and apply supervised fine-tuning, SFT with reasoning and reinforcement learning on it. 
\subsubsection{\textbf{Supervised Fine-Tuning}}
We directly train the base model on \textbf{F}ing\textbf{ER}-\textbf{Instruct}-\textbf{60k}, the response of model only contains "Yes" or "No" answer following the next token prediction paradigm. It means the model only needs to learn predicting the correct answer without any reasoning process. The loss function is Cross-Entropy Loss:
\begin{equation}
\label{loss:logp}
    \mathcal{L}_{CE} = -\sum_{i=1}^{N} \, y_{i} \, \log(p_{i})
\end{equation}
\subsubsection{\textbf{Supervised Fine-Tuning with Reasoning}}
We also train base model on \textbf{F}ing\textbf{ER}-\textbf{Instruct}-\textbf{60k}, but the difference compared to Supervised Fine-Tuning is the model needs to learn predicting the correct answer within $<answer>...</answer>$ tag and its reasoning processes within   $<reason>...</reason>$ tag. We apply prompt engineering on the input tokens to reach this difference. The loss also contains the gap of reasoning processes and the gap of answers.
% , and prompt details are in \textbf{[TODO]}Appendix X.
%
%\textbf{[TODO]} Following the next token prediction paradigm, we adopt the same cross-%entropy loss, as illustrated in Eq.~\ref{loss:logp}. 
% 
%\begin{equation}
%\label{loss:logp}
%    \mathcal{L}_{CE} = -\sum_{i=1}^{N} \, y_{i} \, \log(p_{i})
%\end{equation}
% 

\subsubsection{\textbf{GRPO Training}}
\label{subsec:grpo}
We employ GRPO \cite{shao2024deepseekmath} to enhance reasoning inference performance, exploring two protocols: (i) Zero-GRPO, which relies solely on reinforcement learning without initial supervised data; and (ii) GRPO with cold-start Supervised Fine-Tuning, which combines initial supervised learning with subsequent reinforcement optimization.

% \textbf{Zero-GRPO} 
\paragraph{\textbf{Zero-GRPO}}
% \subsubsection{Zero-GRPO}\textbf{TODO}
Zero-GRPO is an exploratory attempt that is initiated directly from Qwen-2.5-VL \cite{bai2025qwen2} and uses RL to implicitly improve reasoning abilities without annotated reason. For each video-question pair, we first sample a group of outputs $\{o_1, o_2, ...,o_G\}$ by old policy $\pi_{\theta_{old}}(o_i|v,q)$, $v$ denotes the video that needs to be evaluated, $q$ denotes the question for each entity and dimension. Then update the policy model $\pi_\theta$ by minimizing the following loss.
\begin{equation}
\begin{split}
\mathcal{L}_{GRPO}(\theta) &= -\mathbb{E}[q \text{\textasciitilde} P(Q), \{o_i\}_{i=1}^{G} \text{\textasciitilde} {\pi_\theta}_{old}(O|v,q)] 
\\
 &{\frac {1} {G}} \displaystyle\sum_{i=1}^G \Bigg( min\bigg( {\frac {\pi_\theta(o_i|v,q)} {{\pi_\theta}_{old}(o_i|v,q)}}* {Adv}_i, 
 \\
 &clip\Big({\frac {\pi_\theta(o_i|v,q)} {{\pi_\theta}_{old}(o_i|v,q)}},1-{\epsilon}, 1+{\epsilon}\Big) *{Adv}_i \bigg)
 \\
 & + \beta \mathbb{D}_{KL}(\pi_\theta||\pi_{ref}) \Bigg)
\end{split}
\end{equation}

\begin{equation}
\mathbb{D}_{KL}(\pi_\theta||\pi_{ref}) = {\frac {\pi_{ref}(o_i|v,q)} {{\pi_\theta}(o_i|v,q)}} - log{\frac {\pi_{ref}(o_i|v,q)} {{\pi_\theta}(o_i|v,q)}} -1
\end{equation}
$\beta$ denotes the coefficient of Kullback-Leibler Divergence \cite{kullback1951kullback} between base model and policy model, $\epsilon$ denotes the threshold of clip.
$Adv_i$ is the advantage which is the normalization of a group of rewards $\{r_1,r_2,...,r_G\}$ computed from outputs within each group:

\begin{equation}
Adv_i = {\frac {r_i - Mean\{r_1,r_2,...,r_g\}} {Std\{r_1, r_2,...,r_G\}}}
\end{equation}
$r_i$ is composed of two reward functions:

\begin{equation}
r_i = {r_{accuracy}}_i + {r_{format}}_i
\end{equation}
\begin{equation}
{r_{accuracy}}_i = 
\begin{cases}
1.0 &\text{if } answer_i = {GT}_i \\
0.0 &\text{else}
\end{cases}
\end{equation}
\begin{equation}
{r_{format}}_i = 
\begin{cases}
1.0 &\text{if } {o}_i \text{ includes correct format} \\
0.0 &\text{else}
\end{cases}
\end{equation}
Correct format means the output $o_i$ contains two tags: 

$<answer>...</answer>$ and $<reason>...</reason>$.

"Yes" or "No" token only appears within the answer tag, and the reasoning process only appears within reason tag.

% \textbf{GRPO with cold-start Supervised Fine-Tuning}
\paragraph{\textbf{GRPO with cold-start Supervised Fine-Tuning}}
DeepSeek-R1 demonstrated that fine-tuning on an annotated dataset with reasoning processes before applying reinforcement learning (RL) yields better performance than directly using RL \cite{guo2025deepseek}. We adopt this approach in our supervised fine-tuning model. The sole difference between Zero-GRPO and GRPO with cold-start Supervised Fine-Tuning lies in the base model: the latter is initialized from a model pre-trained on annotated data containing reasoning processes.

% \subsubsection{Compute QA Score}
% 
% \begin{equation}
%   logits = P(|)
% \end{equation}
% 
% 
\section{Experiments}
% \subsection{Experiment Settings}
% \subsubsection{Training Details}
% 
\subsection{\textbf{Datasets and Evaluation Metrics}}
\subsubsection{\textbf{Datasets}}
% self-dataset
%In this section, we first introduce the data statistics of our proposed \textbf{F}ing\textbf{ER}-\textbf{test} dataset.

We split 185 generated videos (around $5\%$ of whole data) with 3.5k entity-level questions from 5 distinct quality assessment dimensions to formulate our \textbf{F}ing\textbf{ER}-\textbf{test} dataset.
% public benchmarks
Regarding the public benchmarks, we adopt the popular GenAI-Bench \cite{li2024genai} and recently released MonetBench \cite{xu2024visionreward} for performance evaluation. 
GenAI-Bench contains 800 unique text prompts paired with 4 T2V models, and each generated video has MOS (Mean Opinion Scores) annotated by 3 annotators.
MonetBench consists of 1000 different text prompts, each paired with 2 T2V models. Each pair of videos is generated with the same prompt but different video generation models. MonetBench annotates the video pair with human preferences, including "win", "lose", and "tie" options.
\subsubsection{\textbf{Evaluation Metrics}}
We report the accuracy (Acc) of "Yes" or "No" answers, the Pearson linear correlation coefficient (PLCC), and the Spearman rank correlation coefficient (SRCC) on our proposed FingER-test dataset. 
We evaluate our models with and without token probability calculation, denoted by (\textit{w/o prob}) and (\textit{w/ prob}) in Tab.~\ref{table:zero_shot} and Tab.~\ref{table:sft}.
% \textbf{F}ing\textbf{ER}-\textbf{test} dataset. 
Following previous works in \cite{he2024videoscore, lin2024evaluating}, we utilize the SRCC and the PLCC for evaluating model's performance on GenAI-Bench. 
And we use pairwise accuracy as the metrics for human preference evaluation on MonetBench and report \textit{tau} and \textit{diff}, followed \cite{zhang2024vision, deutsch2023ties}.

\subsection{Implementation Details}
Based on Qwen-2.5-VL-7B \cite{bai2025qwen2}, we fine-tune our model with the following experiment settings: learning rate of 5.0e-6, global batch size of 32, video input fps (frame-per-second) is set to 2, and video maximum input resolution is set to $448 \times 448$ pixels. 
We utilize LLaMA-Factory \cite{zheng2024llamafactory} as our supervised fine-tuning (SFT) codebase.
We perform SFT on our proposed FingER-Instruct-60k dataset for 2 epochs with 8 NVIDIA H20 GPUs, 
and the training steps are the same for the model trained with extra reasoning process. 
% added for supervision.
As for the settings of our reinforcement learning (RL) experiments, we employ Huggingface-TRL \cite{vonwerra2022trl} as our RL fine-tuning tool with following hyper-parameters to implement GRPO: $\beta = 0.04$, and the number of group $G = 16$, $\epsilon = 0.2$, $\mu = 1$, the initial learning rate of RL is 5.0e-7. We train Zero-GRPO and GRPO with cold-start for 2k steps on 4 NVIDIA H20 GPUs.
% setting the coefficient to $0.0$ effectively removes the Kullback-Leibler Divergence from the loss function, a strategy shown to be beneficial in recent studies \cite{chu2025gpg}; 
% $G \in \left\{16 \, , 8 \, , 4\right\}$,
% \textbf{[TODO]}.

\begin{table*}[t]
  \caption{Correlation between model Zero-shot answer and human reference on FingER-test}
  \label{table:zero_shot}
  \resizebox{1.0\linewidth}{!}{
  \begin{tabular}{l|c|c|c|c|c|c}
    \toprule
    % Method (Zero-shot)
    Method & Visual Quality & Temporal & Dynamic Degree & Text Alignment & Factual & Overall \\
    % & GenAI-Bench\cite{li2024genai} \\
    \hline
    Qwen2.5-VL & Acc/SRCC/PLCC & Acc/SRCC/PLCC & Acc/SRCC/PLCC & Acc/SRCC/PLCC & Acc/SRCC/PLCC & Acc/SRCC/PLCC \\
    % \hline
    \midrule
    Overall Level & --- & --- & --- & --- & --- & -/30.68/29.27 \\
    Dimension Level & -/35.06/35.54 & -/16.05/17.06 & -/14.81/14.09 & -/33.68/32.62 & -/13.86/12.28 & -/52.32/61.14 \\
    Entity (w/o prob) & 25.33/1.85/5.22 & 78.72/83.26/\underline{83.91} & 72.87/51.04/48.98 & 81.6/70.68/73.44 & 58.34/\underline{51.03}/\underline{53.27} & 66.50/\underline{80.86}/\underline{83.71} \\
    \textbf{Entity} (w/ prob) & \underline{25.33}/40.60/40.94 & \textbf{78.72}/\underline{84.51}/\textbf{85.44} & \underline{72.87}/\textbf{56.48}/\textbf{56.85} & \textbf{81.6}/\textbf{74.09}/\textbf{76.49} & \textbf{58.34}/\textbf{57.45}/\textbf{58.67} & \textbf{66.50}/\textbf{81.23}/\textbf{85.26} \\
    +Reason (w/o prob) & 45.71/\textbf{49.97}/\underline{49.61} & 77.65/83.12/83.89 & 75.21/\underline{54.30}/\underline{52.87} & 81.08/\underline{73.24}/75.31 & 40.51/17.43/23.55 & 63.96/73.40/79.15 \\
    +\textbf{Reason} (w/ prob) & \textbf{45.71}/\underline{46.29}/\textbf{49.64} & \underline{77.65}/\textbf{84.60}/83.89 & \textbf{75.21}/48.88/52.80 & \underline{81.08}/72.38/\underline{75.35} & \underline{40.51}/29.27/23.50 & \underline{63.96}/73.29/79.18 \\
    % \hline
    % FingER (w/o prob) & /83.48/82.53 & /83.13/83.70 & /71.37/67.95 & /70.94/73.75 & /64.12/64.61 & /88.87/89.67 \\
    % FingER (w/ prob) & /85.31/85.22 & /86.24/86.99 & /77.07/74.73 & /73.85/77.98 & /70.99/69.26 & /90.23/91.41 \\
    \bottomrule
  \end{tabular}
  }
\end{table*}

\begin{table*}[t]
  \caption{Correlation between SFT/RL model answer and human reference on FingER-test (Z-GRPO means Zero-GRPO)}
  \label{table:sft}
  \resizebox{1.0\linewidth}{!}{
  \begin{tabular}{l|c|c|c|c|c|c}
    \toprule
    % Method (Zero-shot)
    Method & Visual Quality & Temporal & Dynamic Degree & Text Alignment & Factual & Overall \\
    % & GenAI-Bench\cite{li2024genai} \\
    \hline
    & Acc/SRCC/PLCC & Acc/SRCC/PLCC & Acc/SRCC/PLCC & Acc/SRCC/PLCC & Acc/SRCC/PLCC & Acc/SRCC/PLCC \\
    % \hline
    \midrule
    % Overall level & - & - & - & - & - & -// \\
    % Dims. level & -// & -// \\
    % Entity (w/o prob) & 25.33/1.85/5.22 & /83.26/83.91 & /51.04/48.98 & 81.6/70.68/73.44 & /51.03/53.27 & /80.86/83.71 \\
    % Entity (w/ prob) & 25.33/40.60/40.94 & /84.51/85.44 & /56.48/56.85 & 81.6/74.09/76.49 & /57.45/58.67 & /81.23/85.26 \\
    % Reasoning (w/o prob) & \\
    % Reasoning (w/ prob) & \\
    % \hline
    GPT-4o \cite{hurst2024gpt} & 62.19/56.24/57.93 & 77.83/78.64/79.13 & 68.31/54.14/57.02 & 83.41/72.20/74.33 & 58.77/48.93/49.51 & 69.92/81.25/82.36 \\
    % Gemini-2.0\cite{team2024gemini} & \\
    % Gemini-Pro\cite{team2024gemini} & \\
    VideoScore \cite{he2024videoscore} & -/22.80/18.55 & -/23.84/26.06 & -/9.49/7.18 & -/19.18/13.87 & -/22.93/18.31 & -/20.39/17.68 \\
    % copy from zero-shot table, row: entity (w/ prob)
    Qwen2.5-VL \cite{bai2025qwen2} & 25.33/40.60/40.94 & 78.72/84.51/85.44 & 72.87/56.48/56.85 & 81.6/74.09/76.49 & 58.34/57.45/58.67 & 66.50/81.23/85.26 \\
    % VideoScore\cite{he2024videoscore} & -/22.80/18.55 & -/23.84/26.06 & -/9.49/7.18 & -/19.18/13.87 & -/22.93/18.31 & -/20.39/17.68 \\
    % VQAScore\cite{} & \\
    \hline
    Z-GRPO (w/o prob) & 76.01/73.39/70.46 & 78.01/83.13/83.82 & 77.93/69.74/68.47 & 84.46/73.80/75.99 & 55.21/47.47/50.33 & 74.51/83.46/86.56 \\
    \textbf{Z-GRPO} (w/ prob) & 76.01/71.83/71.97 & 78.01/81.81/83.86 & 77.93/67.49/68.51 & 84.46/74.38/76.28 & 55.21/42.21/50.15 & 74.51/83.24/86.82 \\
    \hline
    FingER (w/o prob) & 83.78/83.48/82.53 & 83.33/83.13/83.70 & 83.23/71.37/67.95 & 82.77/70.94/73.75 & 72.89/64.12/64.61 & 81.25/88.87/89.67 \\
    \textbf{FingER} (w/ prob) & \underline{83.78}/\textbf{85.31}/\textbf{85.22} & \underline{83.33}/86.24/86.99 & \textbf{83.23}/\textbf{77.07}/\underline{74.73} & 82.77/73.85/77.98 & 72.89/70.99/69.26 & 81.25/90.23/91.41 \\
    % 5e-6
    +Reason (w/o prob) & 84.05/81.51/81.00 & 84.04/85.88/86.63 & 82.49/69.22/68.22 & 86.79/77.87/79.77 & 74.03/67.47/68.41 & 82.33/89.79/91.64 \\
    +\textbf{Reason} (w/ prob) & \textbf{84.05}/\underline{83.85}/\underline{83.87} & \textbf{84.04}/\underline{86.51}/\underline{87.09} & \underline{82.49}/\underline{76.11}/\textbf{76.70} & \textbf{86.79}/\textbf{79.34}/\textbf{83.16} & \underline{74.03}/\underline{71.70}/70.27 & \textbf{82.33}/\underline{90.31}/\underline{92.04} \\
    % Zero-GRPO (w/o prob) & \\
    % Zero-GRPO (w/ prob) & \\
    % 5e-6
    +GRPO (w/o prob) & 82.30/80.62/78.09 & 82.98/85.08/85.57 & 81.63/65.54/64.92 & 85.88/75.74/77.91 & 74.04/68.65/\underline{70.73} & 81.41/89.26/91.25 \\
    +\textbf{GRPO} (w/ prob)  & 82.30/83.76/83.51 & 82.98/\textbf{86.64}/\textbf{87.43} & 81.63/75.05/74.68 & \underline{85.88}/\underline{78.32}/\underline{82.63} & \textbf{74.04}/\textbf{71.87}/\textbf{72.03} & \underline{81.41}/\textbf{90.43}/\textbf{92.41} \\
    % +GRPO (w/ prob) & \textcolor{red}{10} \\
    % \hline
    % Zero-GRPO (w/o prob) & 76.01/73.39/70.46 & 78.01/83.13/83.82 & 77.93/69.74/68.47 & 84.46/73.80/75.99 & 55.21/47.47/50.33 & 74.51/83.46/86.56 \\
    % Z-GRPO (w/o prob) & 76.01/73.39/70.46 & 78.01/83.13/83.82 & 77.93/69.74/68.47 & 84.46/73.80/75.99 & 55.21/47.47/50.33 & 74.51/83.46/86.56 \\
    % Zero-GRPO (w/ prob) & 76.01/71.83/71.97 & 78.01/81.81/83.86 & 77.93/67.49/68.51 & 84.46/74.38/76.28 & 55.21/42.21/50.15 & 74.51/83.24/86.82 \\
    % \textbf{Z-GRPO} (w/ prob) & 76.01/71.83/71.97 & 78.01/81.81/83.86 & 77.93/67.49/68.51 & 84.46/74.38/76.28 & 55.21/42.21/50.15 & 74.51/83.24/86.82 \\
    \bottomrule
  \end{tabular}
  }
\end{table*}

\subsection{Zero-shot Performance on FingER-test}
% 
%In this section, we demonstrate that entity-level video quality assessment is critical to model's understanding of AI-generated videos, through zero-shot evaluation of the base Qwen2.5-VL-7B model with coarse-to-fine assessment granularity on our FingER-test dataset.

We report the zero-shot performance of Qwen2.5-VL across five dimensions on our dataset. Through ablations on resolution, frame rate (fps), and evaluation granularity, we reveal the capabilities of the base model to handle different dimensions, and further demonstrate the crucial importance of integrating entity-level evaluation.

\begin{figure}[t]
  \centering
  % \begin{subfigure}
  \includegraphics[width=0.49\linewidth]{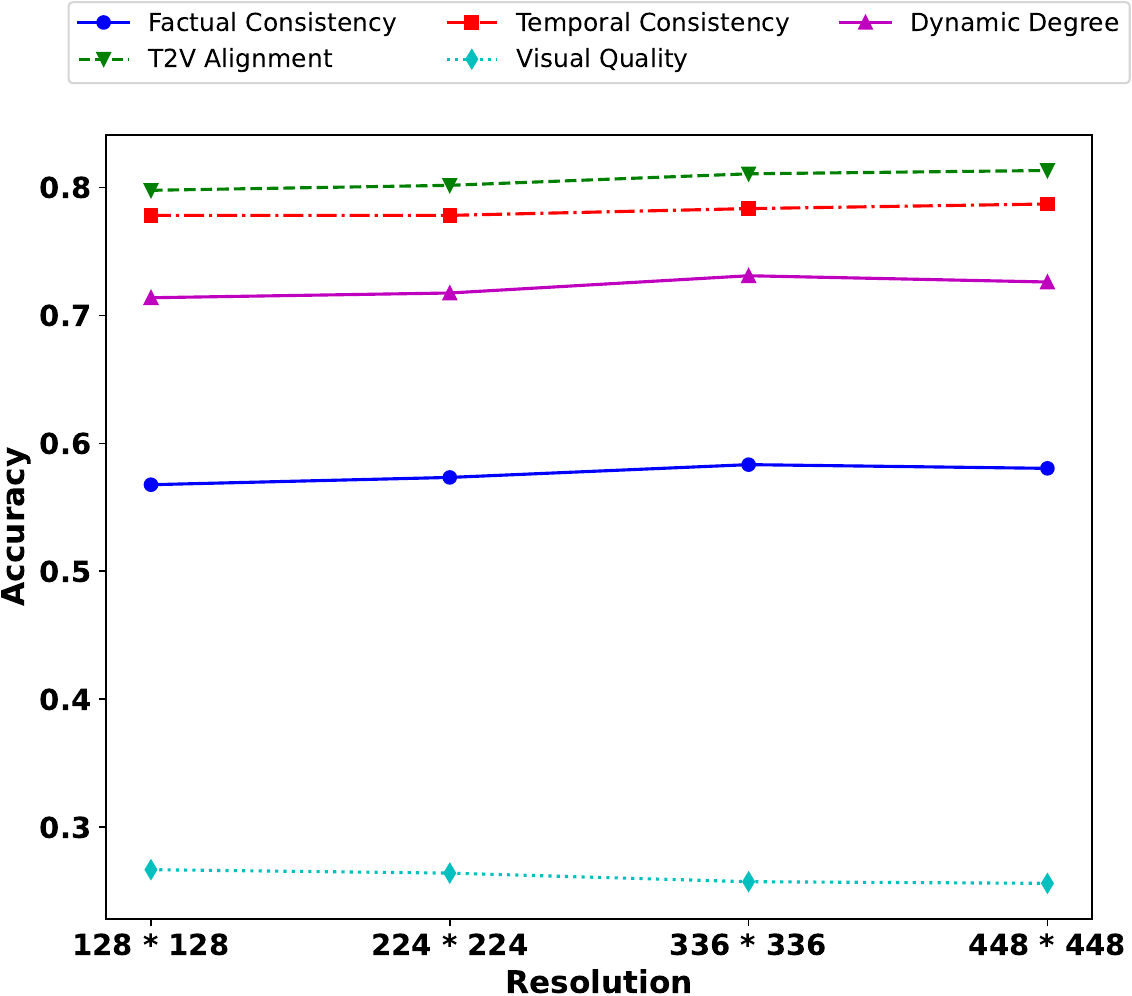}
    % \caption{Subfigure 1}
    % \label{fig:subfig1}
  % \end{subfigure}
  \includegraphics[width=0.49\linewidth]{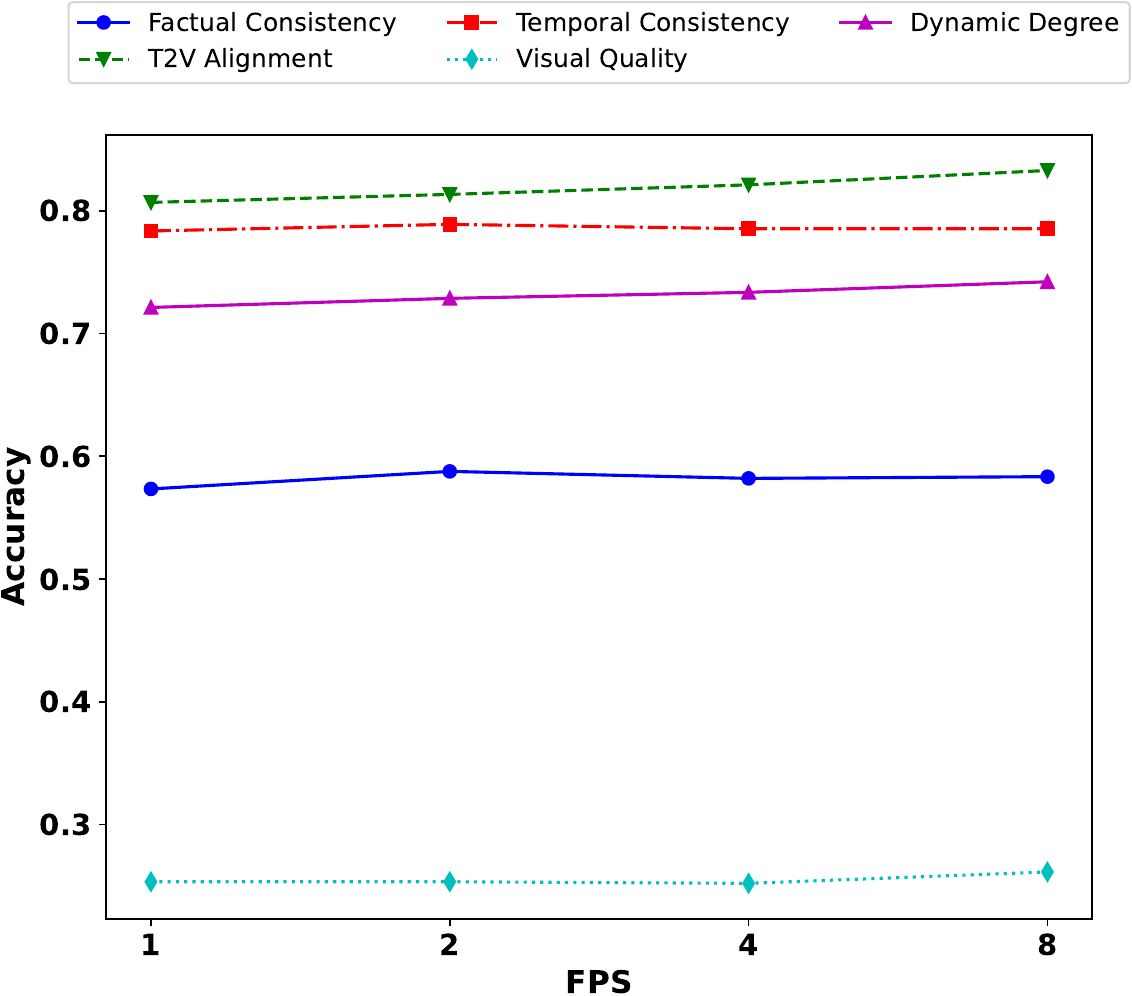}
  \caption{Zero-shot performance on five distinct assessment dimensions with different input resolution and fps.}
  \Description{zero-shot_res_fps}
  \label{fig:res_fps}
\end{figure}

% --- --- --- Lei's instruction
% this analyze the zero-shot performance of qwenvl 2.5 of Accuracy on our dataset, revealing which aspect is hard, what kind of hyper-parameters matters (input size, fps, ) and even what prompt matters. suggest to give a figure.  
% 

\textbf{Increasing resolution and fps leads to slight improvements.} Fig. ~\ref{fig:res_fps} illustrates the accuracy across five dimensions when prompted with entity-level questions. We can see that the accuracy curves show slight improvements with increasing resolutions or frame rates (fps), albeit at a significant computational cost. These results suggest that resolution and fps are not the primary factors of performance enhancement. Consequently, for efficiency we adopt $448 \times 448$ pixels and 2 fps as the default settings for subsequent zero-shot and supervised fine-tuning (SFT) experiments.

\textbf{Performance varies significantly across different dimensions.} As shown in Fig. ~\ref{fig:res_fps}, the zero-shot accuracy for visual quality is exceptionally low at $26.1\%$, while factual consistency achieves $57.6\%$. In contrast, dimensions like text alignment show higher accuracy at $80.59\%$, likely due to the base model's inherent capabilities from pre-training on caption data. We believe that the notably low accuracy in visual quality is primarily attributed to misalignment from AI-generated videos, and the main challenges still lie in dimensions requiring in-depth reasoning, such as factual consistency, temporal consistency, and text alignment, which will be further demonstrated in the following section.

\textbf{Integrating entity-level evaluations brings a substantial performance gain.} To validate the efficacy of our entity-level QA framework, we conduct experiments across three evaluation granularities: \textbf{overall level}, \textbf{dimension level}, and our proposed \textbf{entity level}, as detailed in Tab.~\ref{table:zero_shot}. The \textbf{overall level} (\textit{1st row}) prompts the model with an overall assessment rating from 1 to 4, accompanied by detailed evaluation criteria, while the \textbf{dimension level} (\textit{2nd row}) prompts model to rate each dimension from 1 to 4, which are then averaged to get a final score. The results of our proposed \textbf{entity-level} (\textit{3rd and 4th rows}) are reported with and without a probability calculation strategy introduced in Sec.~\ref{subsec:method_framework}, and furthermore, we instruct the model to provide explanatory reasoning along with answers (\textit{last two rows}). 
Compared to the entity-level framework, both the overall and dimension levels exhibit substantial performance degradation across all dimensions, indicating that fine-grained evaluation substantially enhances the model's performance.
It is worth noting that incorporating explanatory reasoning does not bring improvements, revealing the inherent limitations of the base model in understanding AI-generated videos.

\begin{table}[t]
  \caption{Zero-shot Evaluation Results on Public Benchmarks}
  \label{tab:benchmarks}
  \resizebox{1.0\linewidth}{!}{
  % \begin{tabular}{l|c|cc|cc}
  \begin{tabular}{l|cc|cc}
    \toprule
    % Method & Train data & \multicolumn{2}{c}{GenAI-Bench\cite{li2024genai}} & \multicolumn{2}{c}{MonetBench\cite{xu2024visionreward}} \\
    Method & \multicolumn{2}{c}{GenAI-Bench\cite{li2024genai}} & \multicolumn{2}{c}{MonetBench\cite{xu2024visionreward}} \\
    \hline
    % & & SRCC & PLCC & tau & diff \\
    & SRCC & PLCC & tau & diff \\
    % \hline
    \midrule
    % GPT-4o\cite{hurst2024gpt} & - & - & - & 45.70 & 48.30 \\
    GPT-4o\cite{hurst2024gpt} & 35.79 & 36.61 & 45.70 & 48.30 \\
    % Gemini-Pro\cite{team2024gemini} & - & - & - & 52.20 & 56.80 \\
    % Gemini-Pro\cite{team2024gemini} & - & - & 52.20 & 56.80 \\
    % Qwen2.5-VL\cite{bai2025qwen2} & - & 46.62 & 44.29 & 46.70 & 44.27 \\
    Qwen2.5-VL\cite{bai2025qwen2} & 46.62 & 44.29 & 46.70 & 44.27 \\
    % VideoScore\cite{he2024videoscore} & 37.6k & 42.22 & 40.62 & 49.10 & 54.90 \\
    VideoScore\cite{he2024videoscore} & 42.22 & 40.62 & 49.10 & 54.90 \\
    % VQAScore\cite{lin2024evaluating} &  & 52.70 & 50.60 & 56.10 & 59.50 \\
    VQAScore\cite{lin2024evaluating} & 52.70 & 50.60 & 56.10 & 59.50 \\
    % VQAScore (ECCV 2024)\cite{lin2024evaluating} & 52.7 & 50.6 \\
    % VideoScore\cite{he2024videoscore} & 37.6k & 42.22 & 40.62 & 49.10 & 54.90 \\
    % VisionReward\cite{xu2024visionreward} & - & - & 64.00 & 72.10 \\
    % VideoScore (EMNLP 2024)\cite{he2024videoscore} & & & \\
    % 3k steps, 5.0e-6, w/ reasoning
    % FingER (Zero-shot) & 56.98 & 57.55 \\
    % 4k steps, 5.0e-6, w/o reasoning
    % FingER (Zero-shot) & 3.3k & 56.68 & 57.25 & 57.80 & 62.07 \\
    \hline
    Zero-GRPO & 49.58 & 44.39 & 51.30 & 51.34 \\
    \hline
    % 4k steps, 5.0e-6, w/o reasoning
    % FingER-base & & 54.13 & 52.60 & 53.90 & 57.31\\
    FingER & 54.13 & 52.60 & 53.90 & 57.31 \\
    % 4k steps, 5.0e-6, w/ reasoning
    % + Reasoning, 5e-6 & & \underline{56.68} & \textbf{57.25} & 57.80 & 62.07 \\
    + Reason & \underline{56.68} & \textbf{57.25} & \underline{57.80} & \underline{62.07} \\
    % 3k steps, 1.0e-5, w/ reasoning
    % + Reasoning, 1e-5 & & 57.00 & 57.86 & 58.90 & 63.78 \\
    % without cold start, 200 steps, 5e-8
    % Zero-GRPO, 5e-8 & & 49.58 & 44.39 & - & - \\
    % cold start, 1000 steps
    % + GRPO, 5e-6 & & \textbf{57.03} & \underline{56.59} & 58.00 & 62.80 \\
    + GRPO & \textbf{57.03} & \underline{56.59} & \textbf{58.00} & \textbf{62.80} \\
    % + GRPO, 1e-5 & & & & \\
    % \hline
    % without cold start, 200 steps, 5e-8
    % Zero-GRPO, 5e-8 & & 49.58 & 44.39 & - & - \\
    % Zero-GRPO & 49.58 & 44.39 & 51.30 & 51.34 \\
  \bottomrule
  \end{tabular}
}
\end{table}

\begin{figure*}[t]
  \includegraphics[width=\linewidth]{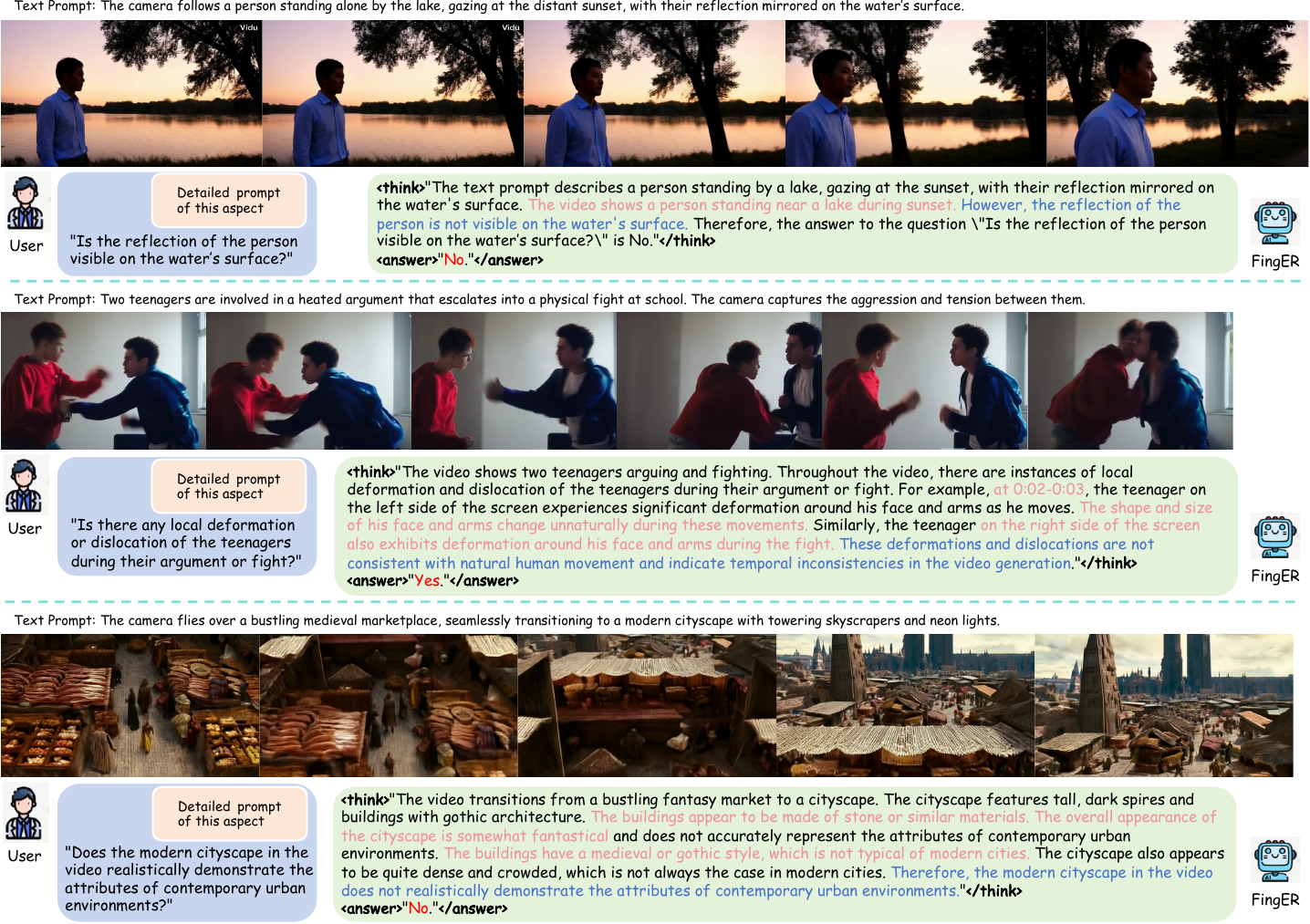}
  \caption{Qualitative results. We show several reasoning results outputted by our GRPO model.}
  \Description{main figure}
  \label{fig:pipeline}
\end{figure*}

\subsection{SFT and RL Performance on FingER-test}
\label{exp:sft_rl}

In this section, we report the performance of our reasoning model on \textbf{F}ing\textbf{ER}-test using different training protocols including SFT with answers, SFT with reasons, zero GRPO, and GRPO with a cold start, we also provide results using the closed-source model GPT-4o and VideScore \cite{he2024videoscore} for comparisons, as detailed in Tab.~\ref{table:sft}. Note that all these results, except for VideoScore \cite{he2024videoscore}, are obtained by entity-level evaluations for fair comparisons.

%In this section, we demonstrate the effectiveness of our proposed \textbf{F}ing\textbf{ER} framework and \textbf{F}ing\textbf{ER}-\textbf{Intruct}-\textbf{60k} dataset, through experimenting with multiple training methods mentioned in Sec.~\ref{subsec:method_training}. 
%First of all, we validate the effectiveness of supervised fine-tuning (SFT) with entity-aware questions and simple "Yes"/"No" answers from five distinct assessment dimensions.

%TODO reformulate and add a picture showing results: 
Our model, trained with only answers, demonstrates significant performance improvements over the base model, achieving overall gains of \textbf{14.75/9.00/6.15} in Acc/SRCC/PLCC, respectively. Substantial improvements are observed in the dimensions of visual quality, dynamic degree, and factual consistency. Note that the improvement in the text alignment dimension is limited, mainly due to its inherent capabilities derived from pre-training data.

Incorporating additional reasoning during training further boosts the performance, particularly in the dimensions of text alignment, factual consistency, and temporal consistency.
For the text alignment dimension, the SFT with reasoning harvests performance gains with \textbf{4.02/5.49/5.18} in Acc/SRCC/PLCC.
These improvements underscore the importance of in-depth video understanding to achieve higher performance in these dimensions.

We further investigate the reasoning training using RL, which includes two kinds of training procedures: (1) Zero-GRPO, and (2) GRPO initialized with a cold-start from reasoning SFT training. The results presented in Table ~\ref{table:sft} reveal that Zero-GRPO fails to predict correct answers. Upon closer examination of the training process, we identified that the issue stems from the reasoning component. Zero-GRPO generates reasons that resemble captions rather than logical reasoning. In contrast, when GRPO is applied with a cold-start initialization from our reasoning SFT model, it is able to surpass the SFT model with only 1k additional training steps. Among these dimensions, we observed steady performance improvements in the \textit{temporal} and \textit{factual consistency} dimensions, with boosts of \textbf{1.15/0.88/2.77} in \textit{factual consistency}. 
We believe that the reasoning cold-start teaches the model to reason in a rough manner, while GRPO guides it towards adopting reasons with correct answers, thereby incentivizing the reasoning capability in the model.

Moreover, we evaluate the performance on our proposed FingER-test dataset with closed-source MLLM \cite{hurst2024gpt} \textit{(1st row)}, and VideoScore \cite{he2024videoscore} \textit{(2nd row)}, our proposed FingER outperforms those methods with a large margin across all five assessment dimensions.

\subsection{Comparison on Public Benchmarks}
Tab.~\ref{tab:benchmarks} demonstrates the consistent improvements achieved by our method on two public benchmarks. We compare our methods with GPT-4o, Qwen2.5-VL and two other approaches. Specifically, with only Yes/No answer prediction, we already outperform all methods on GenAI-Bench, indicating the effectiveness of our fine-grained evaluation framework. Training with reasons and GRPO with a cold-start leads to further improvements with a final 8.21\%/11.83\% SRCC/PLCC relative performance boost. On MonetBench, without any weight fitting, we just average scores of five dimensions, our method is able to achieve 3.39\%/5.55\% relative improvements of tau/diff. It is worth noting that VideoScore \cite{he2024videoscore} is trained using 37.6k training videos, while VQAScore \cite{lin2024evaluating}  utilizes 665k samples, we outperform these methods with only 3.3k training videos without additional training samples from other sources, which is at most one-tenth of the training size adopted by other methods.

\section{Conclusion}
In this paper, we emphasize the critical importance of integrating fine-grained reasoning into AI-generated video quality assessment, and we propose \textbf{F}ing\textbf{ER}, an entity-level fine-grained quality assessment framework with five distinct evaluation dimensions for AI-generated videos. To bridge the gap between non-AI videos and AI-generated videos, we construct a high-quality dataset, \textbf{F}ing\textbf{ER}-\textbf{Instruct}-\textbf{60k}, which consists of 3.3k videos generated by modern T2V models and 60k entity-level question / answering / reasoning pairs. Based on this dataset, we explore multiple training protocols to best incentivize the model's reasoning capability, including reason SFT, zero GRPO and GRPO with a reasoning cold-start. Extensive experiments demonstrate that by utlizing GRPO training with a cold-start, our method not only achieves the best performance on our dataset, but also outperforms other methods and closed-source models on two public benchmarks. And it is worth noting that we achieve SOTA performance with only 3.3k training samples.

\bibliographystyle{ACM-Reference-Format}
\bibliography{sample-base}

% --- --- --- begin appendix --- --- ---
% \newpage
% \appendix
% \newpage
% \section{Appendix}
% \subsection{Dataset Examples}

% \subsection{Prompt for Entity-level Question Generation}









\end{document}